\documentclass[journal]{IEEEtai}

\usepackage[colorlinks,urlcolor=blue,linkcolor=blue,citecolor=blue]{hyperref}

\usepackage{color,array}

\usepackage{graphicx}

\newcommand{\model}{\textsc{nnPRAT}}
\usepackage{amsmath}    % For \min, \max, \| , align, etc.
\usepackage{amssymb}    % For \mathbb, \mathcal, and extra symbols
\usepackage{amsfonts}   % Provides additional math fonts (needed for \mathbb)

\usepackage{graphicx}
\usepackage{nicefrac}
\usepackage{multirow}
\usepackage{algorithm}
\usepackage{algpseudocode}

\usepackage{array}  % For >{\color{red}} syntax
\usepackage{subcaption}
\usepackage{booktabs}
\usepackage{orcidlink}
\usepackage{eso-pic}
\usepackage{lipsum}

\newcommand{\copyrighttext}{%
  \footnotesize{%
    \textbf{\textcopyright{} 2026 IEEE}. Personal use of this material is permitted. Permission from IEEE must be obtained for all other uses, in any current or future media, including reprinting/republishing this material for advertising or promotional purposes, creating new collective works, for resale or redistribution to servers or lists, or reuse of any copyrighted component of this work in other works. DOI: \href{https://doi.org/10.1109/TAI.2026.3681719}{https://doi.org/10.1109/TAI.2026.3681719}%
  }%
}

\AddToShipoutPictureBG*{%
  \AtPageLowerLeft{%
    \put(55,25){%
      \parbox{1.0\textwidth}{%
        \centering\copyrighttext%
      }%
    }%
  }%
}

\newtheorem{lemma}{Lemma}
\begin{document}

\title{Nearest Neighbor Projection Removal Adversarial Training} 

\author{Himanshu Singh$^1$\orcidlink{0009-0000-8946-3674}, \IEEEmembership{Member, IEEE},
A. V. Subramanyam$^1$\orcidlink{0000-0002-8873-4644}, \IEEEmembership{Member, IEEE},
Shivank Rajput$^1$, and \\
Mohan Kankanhalli$^2$\orcidlink{0000-0002-4846-2015}, \IEEEmembership{Fellow, IEEE} \\
$^1$IIIT Delhi, India, $^2$NUS, Singapore}
% \author{First A. Author, \IEEEmembership{Fellow, IEEE}, Second B. Author, and Third C. Author, Jr., \IEEEmembership{Member, IEEE}
% \thanks{This paragraph of the first footnote will contain the date on which you submitted your paper for review. It will also contain support information, including sponsor and financial support acknowledgment. For example, ``This work was supported in part by the U.S. Department of Commerce under Grant BS123456.'' }
% \thanks{The next few paragraphs should contain the authors' current affiliations, including current address and e-mail. For example, F. A. Author is with the National Institute of Standards and Technology, Boulder, CO 80305 USA (e-mail: author@boulder.nist.gov).}
% \thanks{S. B. Author, Jr., was with Rice University, Houston, TX 77005 USA. He is now with the Department of Physics, Colorado State University, Fort Collins, CO 80523 USA (e-mail: author@lamar.colostate.edu).}
% \thanks{T. C. Author is with the Electrical Engineering Department, University of Colorado, Boulder, CO 80309 USA, on leave from the National Research Institute for Metals, Tsukuba, Japan (e-mail: author@nrim.go.jp).}
% \thanks{This paragraph will include the Associate Editor who handled your paper.}}

\markboth{IEEE Transactions on Artificial Intelligence}{Singh \MakeLowercase{\textit{et al.}}: Nearest Neighbor Projection Removal Adversarial Training}
% : Bare Demo of IEEEtai.cls for IEEE Journals of IEEE Transactions on Artificial Intelligence}

\maketitle

\begin{abstract}
Deep neural networks have exhibited impressive performance in image classification tasks but remain vulnerable to adversarial examples. Standard adversarial training enhances robustness but typically fails to explicitly address inter-class feature overlap, a significant contributor to adversarial susceptibility. In this work, we introduce a novel adversarial training framework that actively mitigates inter-class proximity by projecting out inter-class dependencies from adversarial and clean samples in the feature space. Specifically, our approach first identifies the nearest inter-class neighbors for each adversarial sample and subsequently removes projections onto these neighbors to enforce stronger feature separability. Theoretically, we demonstrate that our proposed logits correction reduces the Lipschitz constant of neural networks, thereby lowering the Rademacher complexity, which directly contributes to improved generalization and robustness. Extensive experiments across standard benchmarks including CIFAR-10, CIFAR-100, SVHN, and TinyImagenet show that our method demonstrates strong performance that is competitive with leading adversarial training techniques, highlighting significant achievements in both robust and clean accuracy. Our findings reveal the importance of addressing inter-class feature proximity explicitly to bolster adversarial robustness in DNNs. The code is available in the supplementary material.

\end{abstract}

\begin{IEEEImpStatement}
This work advances adversarial robustness by introducing a theoretically grounded training framework that explicitly removes inter-class feature projections. Our method enforces geometric separability in representation space, reducing inter-class entanglement, a key yet underexplored cause of adversarial vulnerability. The proposed projection removal operation lowers the Lipschitz constant and Rademacher complexity of the network, providing formal guarantees of improved generalization and stability. Our approach enhances robustness with negligible computational overhead. By bridging geometric feature disentanglement with adversarial training, this work offers a new direction for building models that are simultaneously accurate, theoretically interpretable, and resilient to adversarial manipulation. The ideas herein can generalize to other safety critical domains requiring feature level robustness and stability.
\end{IEEEImpStatement}


\section{Introduction}
\label{sec:intro}
Deep neural networks (DNNs) have become \textit{de-facto} decision-making engines in safety critical domains, including autonomous driving and medical imaging \cite{zech2018variable, o2022network, bi2025lane}. Their ability to learn complex patterns from large-scale data has enabled unprecedented breakthroughs in tasks such as object detection, semantic segmentation, and disease classification. Despite their impressive performance, DNNs have a well-documented vulnerability in which imperceptible yet malicious \emph{adversarial perturbations} may generate erroneous and potentially catastrophic predictions \cite{szegedy2013intriguing, madry2018towards}. As a result, understanding and mitigating such vulnerability has emerged as a key research area in trustworthy machine learning and computer vision. The mainstream defence paradigm is \emph{adversarial training}, which augments optimisation with worst case perturbed instances so that the learned decision boundary is locally insensitive to prescribed $\ell_p$ bounded attacks \cite{madry2018towards}. State-of-the-art variants such as MART \cite{Wang2020Improving}, squeeze-training (ST) \cite{li2023squeeze}, AR-AT \cite{waseda2025rethinking} and DWL-SAT \cite{xu2025dynamic} substantially improve robustness by balancing clean accuracy and a surrogate of robust risk. 

Despite the significant progress made by recent adversarial defense systems, current approaches have the following limitations: \textbf{(i)}~They predominantly treat robustness as a point-wise phenomenon, ignoring how inter-class feature entanglement in representation space influence models to adversarial attacks \cite{madry2018towards, mustafa2019hidden}. As a result, even adversarially trained networks frequently learn overlapping class representations, which an attacker may exploit using low-cost perturbations. \textbf{(ii)}~Existing formulations offer limited theoretical insight into how the geometry of the last-layer embedding influences generalisation under attack. As a result, improvements are often driven by heuristic regularizers whose impact on model complexity remains poorly understood \cite{li2023squeeze,jiang2025improving}. We address these gaps by revisiting the role of feature geometry in adversarial robustness. Specifically, we observe that one reason for failure is the projection of a sample onto the span of its nearest inter-class neighbor in the feature space. If this projection is not controlled, a small input-space perturbation can move the representation across the decision boundary even when the classifier has been adversarially trained. Building on this, we propose \emph{Nearest Neighbor Projection Removal Adversarial Training} (\model). At each iteration, \model\ first identifies the closest sample from a competing class in the current feature space. It then removes the component of the adversarial (and clean) feature that is aligned with this nearest competitor before the loss is computed. Analytically, we show that the resulting logits correction shrinks the spectral norm of the final linear map, and lowers the Rademacher complexity of the model. Empirically, integrating projection removal into adversarial training yields consistent gains in robust accuracy on CIFAR-10 and CIFAR-100. In summary, we contribute to the field of adversarial robustness in following ways: 
% \begin{itemize}
%     \item First, it illuminates inter-class projection as a previously overlooked mechanism that amplifies adversarial vulnerability. 
%     \item Second, it introduces a theoretically grounded correction that can be seamlessly inserted into off-the-shelf adversarial training pipelines. 
%     \item Third, it demonstrates through extensive experiments that nnPRAT yields robustness and clean accuracy competitive with the latest adversarial defenses. 
% \end{itemize}

\begin{itemize}
    \item We identify inter-class projection as a key component of adversarial vulnerability in neural networks. We show that this projection significantly increases the likelihood of misclassification under attack, by analyzing how features from different classes interact in the latent space. 
    \item We propose, \model, a theoretically grounded correction mechanism that directly mitigates inter-class projection. This approach is lightweight and model-agnostic, making it easy to plug into existing adversarial training pipelines without heavy computational overhead.
    \item We validate our approach through extensive experiments across multiple benchmarks, showing that \model\ consistently improves both robustness and clean accuracy. 
\end{itemize}
By explicitly disentangling class features during training, our method provides a principled approach towards building DNNs that are both accurate and resilient to adversarial manipulation.

\section{Related Works}
\label{sec:background}
% Deep neural networks (DNNs) have achieved remarkable success in a wide array of visual recognition tasks, revolutionizing fields from autonomous driving to medical imaging. However, their susceptibility to adversarial examples small, often imperceptible perturbations that mislead classifiers, remains a critical barrier to their deployment in safety-critical applications. Over the past few years, the machine learning community has increasingly seen adversarial training as the most effective and reliable defense mechanism against such vulnerabilities. This section traces the conceptual and algorithmic progression of adversarial training, starting with the robust optimization framework pioneered by Madry \emph{et al.}~\cite{madry2018towards} and advancing through innovations in loss design, landscape regularization, and feature-space geometry. 
In this section, we review the adversarial training methods. The seminal work of Madry \emph{et al.}~\cite{madry2018towards} formalized adversarial defense as a saddle-point optimization problem, expressed as:  
\[  
\min_{\theta}\;\mathbb{E}_{(x,y)\sim \mathcal{D}}\;\max_{\|\,\delta\|_p\le \epsilon}\!\! \ell\bigl(f_\theta(x+\delta),y\bigr),  
\]  
where the inner maximization seeks the worst-case perturbation within an $\epsilon$-bounded $p$-norm ball, and the outer minimization trains the model parameters $\theta$ to mitigate this adversarial loss. They proposed multi-step projected gradient descent (PGD) as a practical first-order method for solving the inner maximization. Their extensive experiments on datasets like MNIST and CIFAR-10 uncovered two pivotal insights, first, a sufficiently strong first-order adversary, such as PGD, can approximate near worst case perturbations without requiring higher order methods and  second, optimizing for worst case loss significantly enhances robustness but often at the expense of standard (clean) accuracy. Subsequent theoretical analyses, notably by Tsipras \emph{et al.}~\cite{tsipras2018robustness}, provided rigorous evidence that this trade-off between accuracy and robustness may be inherent to certain data distributions, particularly when robust and non robust features conflict. This realization shifted the research focus from maximizing robustness in isolation to achieving a balanced compromise between robustness and generalization.  
  
Building on the foundational insights of PGD-based adversarial training, Zhang \emph{et al.}~\cite{Zhang2019theoretically} introduced TRADES, a method that explicitly decomposes the robust risk into two components, the natural classification error on unperturbed inputs and a boundary error capturing the probability mass near the decision boundary within an $\epsilon$-ball. By substituting the discontinuous indicator function with a Kullback-Leibler (KL) divergence, TRADES formulates the objective as:  
\[  
\sum\nolimits_{i}\!\Bigl[\ell\bigl(f_\theta(x_i),y_i\bigr) + \beta\,\max_{\|\delta\|\le \epsilon} KL\!\bigl(f_\theta(x_i)\,\|\,f_\theta(x_i+\delta)\bigr)\Bigr],  
\]  
where the hyperparameter $\beta$ directly controls the trade-off between clean accuracy and robustness. Notably, the label-agnostic nature of the KL regularizer facilitated semi-supervised extensions, such as Robust Self-Training (RST) by Carmon \emph{et al.}~\cite{carmon2019unlabeled}, which harnesses large volumes of unlabeled data to further narrow the accuracy gap between robust and standard models, demonstrating the potential of data augmentation in robust learning.  
  
While TRADES applies uniform regularization across all samples, subsequent methods recognized the importance of tailoring optimization to specific sample characteristics. Misclassification-Aware Adversarial Training (MART)~\cite{Wang2020Improving} distinguishes between correctly and incorrectly classified samples, augmenting a TRADES-style loss with an additional margin penalty exclusively for benign inputs that are already misclassified. This targeted approach prioritizes optimization effort on hard examples. These results underscore the critical role of the misclassified sample distribution in shaping robust learning outcomes and highlight the value of adaptive loss designs that respond to individual sample difficulties rather than applying a one-size-fits-all regularization. On similar lines, DWL-SAT \cite{xu2025dynamic} first computes a robust distance for each sample with the FAB \cite{croce2019minimally} attack, labelling examples near the decision boundary as fragile. It then converts these distances into exponential weights that boost gradients on vulnerable points and suppress them on already-robust ones. Finally, it embeds the weights into a TRADES-style loss. 
  
Empirical observations have consistently shown that robust models tend to reside in flatter regions of the loss landscape compared to their standard counterparts, which often converge to sharp minima prone to overfitting. Adversarial Weight Perturbation (AWP)~\cite{wu2020adversarial} implemented this insight by introducing a dual perturbation strategy. AWP perturbs model weights in the direction that maximizes loss increase before performing a descent update. This process fosters solutions that are resilient to both data and parameter noise, effectively combating the phenomenon of robust overfitting, where robust accuracy peaks early in training and subsequently declines. When integrated with frameworks like TRADES, AWP establishes a robust baseline, against AutoAttack on CIFAR-10 without requiring additional data, thus illustrating the power of landscape-flattening techniques in enhancing model stability.  
  
Traditional adversarial training methods predominantly focus on high-loss adversarial directions, targeting the peaks of the loss landscape. In contrast, Li \emph{et al.}~\cite{li2023squeeze} propose an innovative perspective with collaborative examples, perturbations that decrease the loss, thereby exploring the valleys of the loss surface. Their ST framework regularizes both the maximal (adversarial) and minimal (collaborative) divergence within each $\epsilon$-ball, penalizing the disparity between adversarial and collaborative neighbors. When combined with techniques like AWP or RST, squeeze training achieves state-of-the-art performance.  
  
Beyond loss landscape modifications, recent efforts have explored the representational properties of neural networks as a means to address adversarial vulnerabilities. Methods focusing on feature-space geometry aim to enhance robustness by increasing inter-class separation in the learned feature representations. These approaches often involve manipulating the feature vectors to reduce overlap between classes, thereby making it harder for small perturbations to cross decision boundaries. Such strategies target the underlying structure of the data representations, complementing input-space and loss-based defenses by addressing adversarial susceptibility at a deeper, model-intrinsic level.  

ARREST \cite{satoshi2023adversarial} mitigates the accuracy–robustness trade-off by adversarially finetuning a clean pretrained model while preserving latent representations. Representation guided distillation and noisy replay prevent harmful representation drift. Building on this representation centric approach, Asymmetric Representation–regularised Adversarial Training (AR-AT) \cite{waseda2025rethinking} introduces a one-sided invariance penalty. The penalty is applied exclusively to adversarial features. This design significantly improves clean accuracy on CIFAR-10 without sacrificing robustness. As a result, AR-AT decisively enhances the accuracy–robustness trade-off that has long been regarded as a fundamental limitation of adversarial training. Kuang \emph{et al.} \cite{kuang2023scarl} looks at semantic information, revealing that adversarial attacks disrupt the alignment between visual representations and semantic word representations. The authors proposed SCARL framework that integrates semantic constraints into adversarial training by maximizing mutual information and preserving semantic structure in the representation space. A differentiable lower bound facilitates efficient optimization. Complementing this line of work, Self-Knowledge-Guided Fast Adversarial Training (SKG--FAT) \cite{jiang2025improving} revisits training on single step FGSM examples and demonstrates that a combination of class-wise feature alignment and relaxed label smoothing can improve robustness while completing training within one GPU-hour. 
  
These contributions collectively illustrate an emerging consensus. Imposing carefully targeted regularisers in feature space or parameter space, can substantially elevate clean performance. They can also reduce computational overhead without compromising adversarial robustness.
Our projection removal adversarial training follows the same philosophy. It achieves class separation by explicitly excising inter-class projections from deep features. This mechanism is orthogonal to the invariance, self-distillation, and weight-perturbation strategies mentioned above.

\section{Methodology}
\label{sec:methodology}
In this section, we present the details of Nearest Neighbor Projection Removal Adversarial Training (\model). We begin by describing the full training algorithm, accompanied by pseudocode, then develop a theoretical analysis that motivates our projection‐removal operation. We also illustrate its geometric effect on a toy example.  

\subsection{Motivation}
Learning-based defenses often fail because adversarial perturbations exploit \emph{high-curvature, low-margin} directions. These directions align closely with class-conditional logit axes in feature space, yet remain almost invisible in pixel space \cite{goodfellow2015explaining, Ilyas2018, Fawzi2018Adversarial}. Adversarial training methods try to blunt this effect by embedding projected gradient steps into every mini-batch \cite{madry2018towards, gowal2020uncovering}. However, the extra steps inflate computational cost and can degrade clean accuracy \cite{rice2020overfitting}.
% Learning‐based defenses often fail because adversarial perturbations exploit \emph{high curvature, low margin} directions that are strongly aligned with class-conditional logit directions in feature space, while nearly invisible in pixel space \cite{goodfellow2015explaining, Ilyas2018, Fawzi2018Adversarial}. Adversarial training methods try to blunt this effect by embedding projected gradient steps into every mini-batch \cite{madry2018towards, gowal2020uncovering}, but it inflates computation and can even degrade clean accuracy \cite{rice2020overfitting}. 

\begin{figure}
    \centering
    \includegraphics[trim=0.0in 0.0in 8.0in 0.0in, clip, width=\columnwidth]{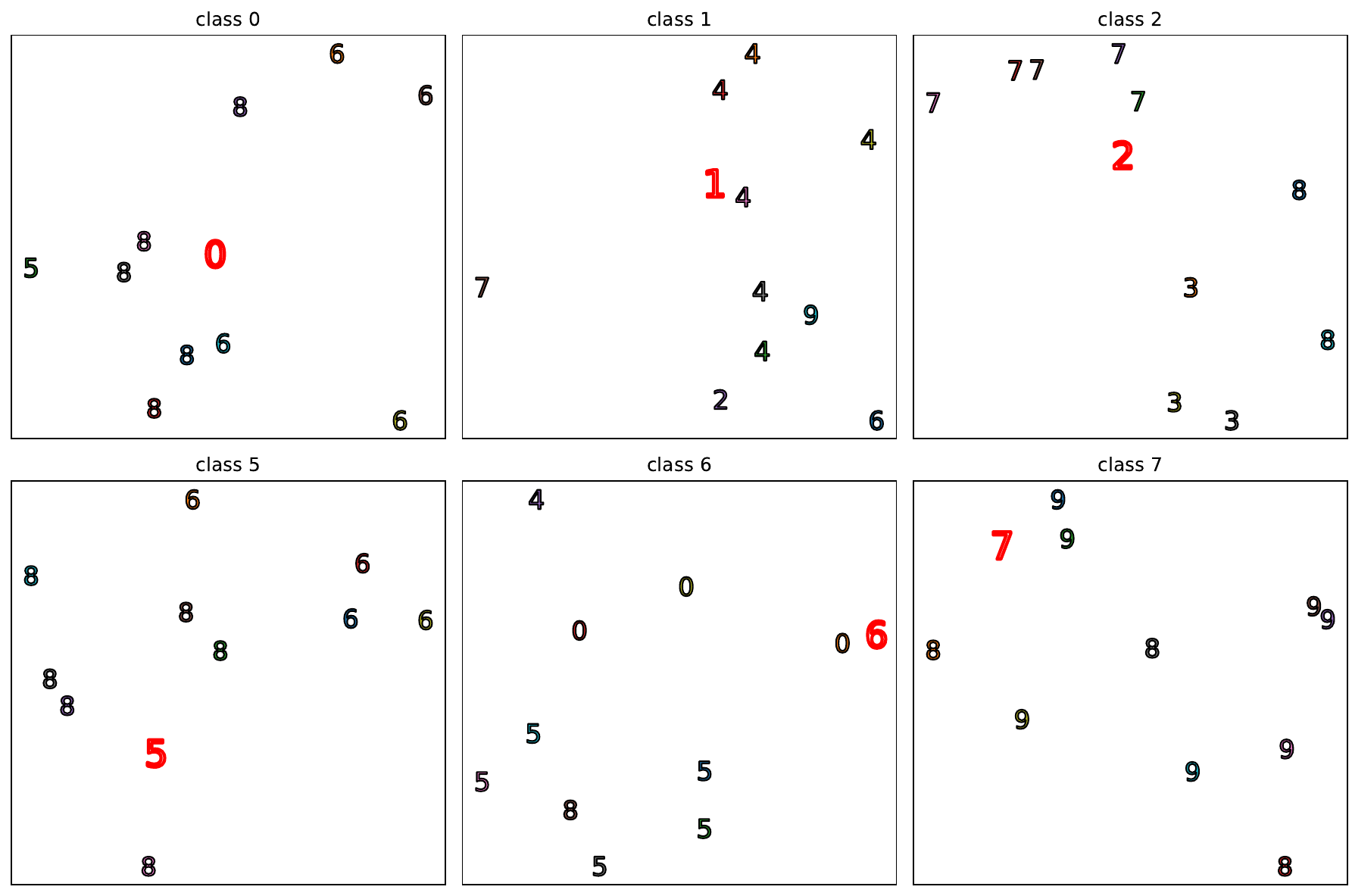}
    \caption{Visualization of the PCA-reduced feature space from a FGSM-trained MNIST model. The red digits indicate the query points, while the other blue digits represent their top-10 nearest neighbors from various classes. Despite adversarial training, queries are majorly surrounded by single off-class neighbors, indicating persistent inter-class entanglement in the representation.
    % Nearest inter-class neighbors of randomly selected samples from each MNIST class.
    }
    \label{fig:motivation}
\end{figure}

Despite its success in reducing worst-case error, first-order adversarial training often produces feature representations that remain insufficiently disentangled. Distinct class manifolds can still develop narrow bridges within the embedding space. Adversarial perturbations readily exploit these bridges \cite{fawzi2018analysis, shamir2022dimpledmanifol}. To characterize this phenomenon, we examine the  penultimate layer features of an FGSM-trained MNIST classifier. We first reduce the features to two dimensions via PCA. For each query point, we then retrieve its top‑k inter-class nearest neighbors. Figure \ref{fig:motivation} visualizes 10 representative query points alongside their $k$ nearest inter-class neighbors ($k$ = 10). Notably, each query point is surrounded almost exclusively by points from a single inter-class. For example, class 5 query draws neighbors primarily from class 8. Even after adversarial training the nearest neighbors in feature space often originate from other classes. This reveals that adversarial training largely enforces local flatness without guaranteeing large angular or Euclidean margins between classes \cite{tsipras2018robustness}. This persistent inter-class entanglement motivates our proposed nearest-neighbor dispersion approach, which explicitly penalizes proximity to off-class embeddings and thereby seeks to complement flatness-based defenses with geometry-aware margin maximization.

For each sample, our \emph{projection-removal} step subtracts the logit vector that points toward the nearest \emph{inter-class} neighbor. Projection removal pushes the corrected logits away from those neighboring logits, which in turn strengthens robustness. This effectively removes the shared, attack susceptible subspace identified by Zhang \emph{et al.} \cite{zhang2019defense} and Carlini \& Wagner \cite{carlini2017towards}. This reduces its spectral norm and hence the product of layer Lipschitz constants, a quantity that controls both adversarial vulnerability \cite{cisse2017parseval, yoshida2017spectralnormregularizationimproving} and PAC-Bayes generalisation bounds \cite{bartlett2017spectrally}.
% The result is a network whose capacity contracts only in the directions attackers prefer, while decision relevant directions remain intact.

% \begin{figure}[htbp]
%     \centering
%     \includegraphics[width=1.0\linewidth]{wacv2026/figures/toy.pdf}
%     \caption{Illustration on a toy 2D Gaussian dataset showing the effect of projection removal in the feature space.  %Effect of projection removal on the feature space.
%     }
%     \label{fig:projection_removal}
% \end{figure}

% Figure \ref{fig:projection_removal} (toy 2D Gaussian dataset) visualises the geometric pay-off. The dashed curve is the original boundary learned with cross-entropy. The red circles mark adversarial examples found by a 10 step PGD attack. Blue crosses are the \emph{projection removed} counterparts of the red points, all now lie safely on the correct side. The boundary is also noticeably flatter, echoing the reduction in Rademacher complexity established in Theorem \ref{thm:complexity} and empirically linked to better clean-accuracy/robustness trade-offs on various datasets in Section \ref{sec:results}. This minimalist, plug-and-play correction therefore offers a computationally light yet theoretically principled path to stronger, margin-aware decision surfaces.

\subsection{Projection Removal}
Motivated by the observation that most misclassifications originate from inter-class entanglement in a highly non-flat loss landscape, we propose to explicitly decouple class features by removing the projection of every example onto its nearest inter-class neighbor. We employ the widely-used Projected Gradient Descent (PGD) algorithm for generating adversarial perturbations. Given a clean input sample $x$, an adversarially perturbed sample $x_{adv}$ is generated using the following update rule:
\begin{equation}
x^{t+1} = \Pi_{B_{\epsilon}[x]}\left(x^{t} - \alpha \cdot \text{sign}(\nabla_{x^{t}} \mathcal{L}(f_\theta(x^{t}), y))\right)
,
\end{equation}
where $\epsilon$ controls the maximum perturbation magnitude, $\alpha$ is the step size, $\mathcal{L}$ denotes the cross-entropy (CE) loss, $f_\theta$ is the neural network classifier parameterized by weights $\theta$, and $y$ is the true label of the input.

To explicitly address inter-class confusion, we identify the nearest neighbor belonging to a different class within the feature representation space. Given an adversarially perturbed example $x_{adv}$, we determine the closest inter-class sample $x_j^*$ based on the Euclidean distance in the feature representation $z = f_\theta(x)$:
\begin{equation}
z_j^* = \underset{j}{\text{argmin}} \|z_{adv} - z_j\|_2, \quad \text{subject to} \quad y_j \neq y_{adv}.
\end{equation}
To strengthen class separability, we remove the projection of the closest inter-class sample from the adversarial example. The projection removal is mathematically defined as:
\begin{equation}
\tilde{z}_{adv} = z_{adv} - \lambda \frac{\langle z_{adv}, z_j^* \rangle}{\lVert z_{adv}\rVert^2} z_{adv},
\end{equation}
where $\lambda$ is a hyperparameter that determines the intensity of projection removal. The projection strength $\lambda$ governs a trade-off between inter-class separation and intra-class compactness. Moderate values suppress shared inter-class directions while preserving class-specific variance, whereas excessively large $\lambda$ may over-attenuate dominant features and weaken intra-class compactness. This removal operation is similarly applied to the clean samples for consistent feature refinement.

The training of the neural network parameters incorporates a combined loss that integrates adversarially refined samples and their clean counterparts, effectively balancing robustness with generalization:
\begin{equation}
\mathcal{L}_{adv} = \mathcal{L}(\tilde{z}_{adv}, y) + \beta\mathcal{L}(\tilde{z}, y).
\end{equation}
Optimizing the joint loss simultaneously enforce class separability and improves robustness. The implementation is given in Algorithm \ref{alg:projection_removal}.

\begin{algorithm}[ht]
\caption{Nearest Neighbor Projection Removal Adversarial Training}
\label{alg:projection_removal}
\begin{algorithmic}[1]
\Require Dataset $X, Y$, neural network $f_{\theta}(x)$
\Require Hyperparameters: $\lambda, \epsilon, \eta, \alpha, \beta$
\Ensure Robust trained model $f_{\theta}(x)$

\State Initialize network parameters $\theta$
\For{$epoch = 1, \dots, M$}
    \For{each batch $(x, y)$}
        \State $x_{adv} = \Pi_{B_{\epsilon}[x]}\left(x^{t} - \alpha \cdot \text{sign}(\nabla_{x^{t}} \mathcal{L}(f_\theta(x^{t}), y))\right)$
        % \State $x_{\text{adv}} = x + \epsilon \cdot \text{sign}(\nabla_x \mathcal{L}(f_{\theta}(x), y))$
        \State $z_j^* = \arg\min_{y_j \neq y_{adv}} \|z_{\text{adv}} - z_j\|_2$
        \State $\tilde{z}_{\text{adv}} = z_{\text{adv}} - \lambda \frac{\langle z_{\text{adv}}, z_j^* \rangle}{\|z_{\text{adv}}\|_2} z_{\text{adv}}$
        \State $\tilde{z} = z - \lambda \frac{\langle z, z_j^* \rangle}{\|z\|_2} z$
        \State $\mathcal{L}_{\text{adv}} = \mathcal{L}(\tilde{z}_{\text{adv}}, y) + \beta\mathcal{L}(\tilde{z}, y)$
        \State $\theta \leftarrow \theta - \eta \nabla_{\theta} \mathcal{L}_{\text{adv}}$
    \EndFor
\EndFor
\State \Return robust trained model $f_{\theta}(x)$
\end{algorithmic}
\end{algorithm}  
   
By integrating projection removal into adversarial training, \model\ explicitly counters inter‐class confusion. Importantly, this drives  the model to push the projection stripped variants away from the decision boundary, pulling samples of the same class closer together and expanding the separation between different classes.
\subsection{Theoretical Analysis}
\label{sec:theory}

\noindent\textbf{Notations.}  
Let $h_\theta:\mathbb{R}^d\!\to\!\mathbb{R}^m$ be the penultimate representation,  
$W_r\!\in\!\mathbb{R}^{C\times m}$ be the weights and $z=W_rh_\theta(x)$ the logits, \emph{C} be the number of the classes.  
For any matrix $A$, $\|A\|_{\mathrm{op}}$ denotes its spectral norm.

% \vspace{0.3em}
\paragraph{Inter-class Projection Removal.}  
Given the nearest-neighbor logits $\tilde{z}$ from a \emph{different} class,  
we remove their projection from $z$:

\begin{equation}
z^{\ast}=z-\frac{z^{\top}\tilde{z}}{\|z\|^{2}}\,z .
\label{eq:zstar}
\end{equation}
This operation reduces  the last layer' Lipschitz constant, as we quantify next.

% \vspace{0.3em}
% \paragraph{Complexity via Covering Numbers.}  
% Rademacher complexity obeys the Dudley entropy integral \cite{mohri2012foundations}

% \begin{equation}
% \label{eq:dudley}
% \begin{aligned}
% R_n(\mathcal{F})
% \le
% \mathcal{O}\!\Bigl(
% \inf_{\varepsilon>0}\Bigl[
% \varepsilon\\[3pt]
% &+\frac{1}{\sqrt{n}}\int_{\varepsilon}^{\sqrt{n}}
% \sqrt{\log N\bigl(\delta,\mathcal{F},\|\cdot\|)}\,
% \mathrm{d}\delta
% \Bigr]
% \Bigr).
% \end{aligned}
% \end{equation}
% Thus, any contraction of $\log N(\cdot)$ tightens generalisation bounds.

% \vspace{0.3em}

% \vspace{0.5em}
\begin{lemma}\label{lem:lipschitz}
Let $z$ and $\tilde{z}$ be the sample and nearest neighbor's logits. Then the projection removal step induces a spectral norm contraction given by $\|W_r'\|_{\mathrm{op}}
\;\le\;(1-\alpha)\,\|W_r\|_{\mathrm{op}}$, where $\alpha \in (0,1)$. 
\end{lemma}

\noindent \textbf{Proof.} The projection removal can be written as,
\begin{equation}
z' \;=\;\Bigl(1-\alpha\,\dfrac{z^{\top}\tilde{z}}{\|z\|^{2}}\Bigr){z}.
\label{eq:matrix}
\end{equation}

Since $z = W_rh_{\theta}(x)$, we can write,
\begin{equation}
    z' = \Bigl(1 - \alpha \frac{z^\top\tilde{z}}{\|z\|^2}\Bigr)z = W_r'h_{\theta}(x).
\end{equation}
The modified last-layer weight matrix becomes:
\begin{equation}
 W_r' = \Bigl(1 - \alpha \frac{z^\top\tilde{z}}{\|z\|^2}\Bigr)W_r.
\end{equation}
The Lipschitz constant of this layer is given by, $L = \|W_r\|_{\mathrm{op}}$ \cite{gouk2021regularisation}.

After correction, the new Lipschitz constant is:
\begin{equation}
 L' = \|W_r'\|_{\mathrm{op}} = \| (1 - \alpha \frac{z^\top\tilde{z}}{\|z\|^2})W_r\|_{\mathrm{op}}.
\end{equation}
Thus, the new Lipschitz constant satisfies:
\begin{equation}
 L' = (1-\alpha \frac{z^\top\tilde{z}}{\|z\|^2})L. 
\end{equation}
Since $z$ and $\tilde{z}$ are closest neighbors, their similarity is high. Thus, $\Bigl(1-\alpha \frac{z^\top\tilde{z}}{\|z\|^2}\Bigr) \approx 1-\alpha < 1$, which implies,
\begin{equation}
 L' < L. 
 \label{eq:lipschitz}
\end{equation}
% \vspace{0.5em}
\begin{lemma}\label{lem:complexity}
Let $\mathcal{F}'$ be the network class obtained by applying \eqref{eq:zstar} (or equivalently \eqref{eq:matrix}) to every logit vector. Let $\mathcal{R}_n(F)$ be the Rademacher complexity of $\mathcal{F}$. Then the Rademacher complexity of $\mathcal{R}_n(F')$ holds, $\mathcal{R}_n(\mathcal{F}') \leq (1-\alpha) \mathcal{R}_n(\mathcal{F})$.
% Then, for any $\varepsilon>0$,
% \[
% \log N\!\bigl(\varepsilon,\mathcal{F}',\|\cdot\|_\infty\bigr)
% \;\le\;
% \log N\!\bigl((1-\alpha)^{-1}\varepsilon,\mathcal{F},\|\cdot\|_\infty\bigr),
% \]
% and consequently, by \eqref{eq:dudley},
% \(
% R_{n}(\mathcal{F}')\le(1-\alpha)\,R_{n}(\mathcal{F}).
% \)
\end{lemma}

Since $W_r'$ directly contributes to the Lipschitz constant of the network, a reduction in its Lipschitz constant also reduces the Rademacher complexity. \textcolor{black}{Following \cite{yin2019rademacher, xiao2024bridging}, the adversarial setting admits bounds in terms of Rademacher complexity. Thus reducing this complexity tightens robust generalization bounds, which we target with our regularization.}

\begin{figure*}
  \centering
  \begin{subfigure}{0.32\linewidth}
     \centering
    \includegraphics[width=\linewidth]{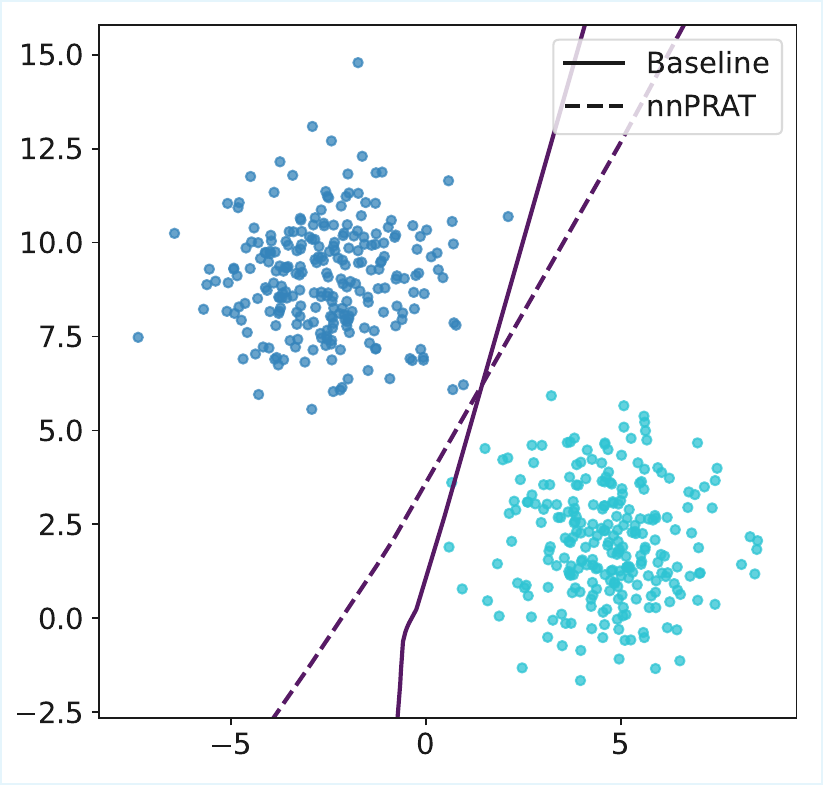}
    \caption{}
    \label{fig:projection_removal}
  \end{subfigure}
  \begin{subfigure}{0.31\linewidth}
    \centering
    \includegraphics[width=1.0\linewidth]{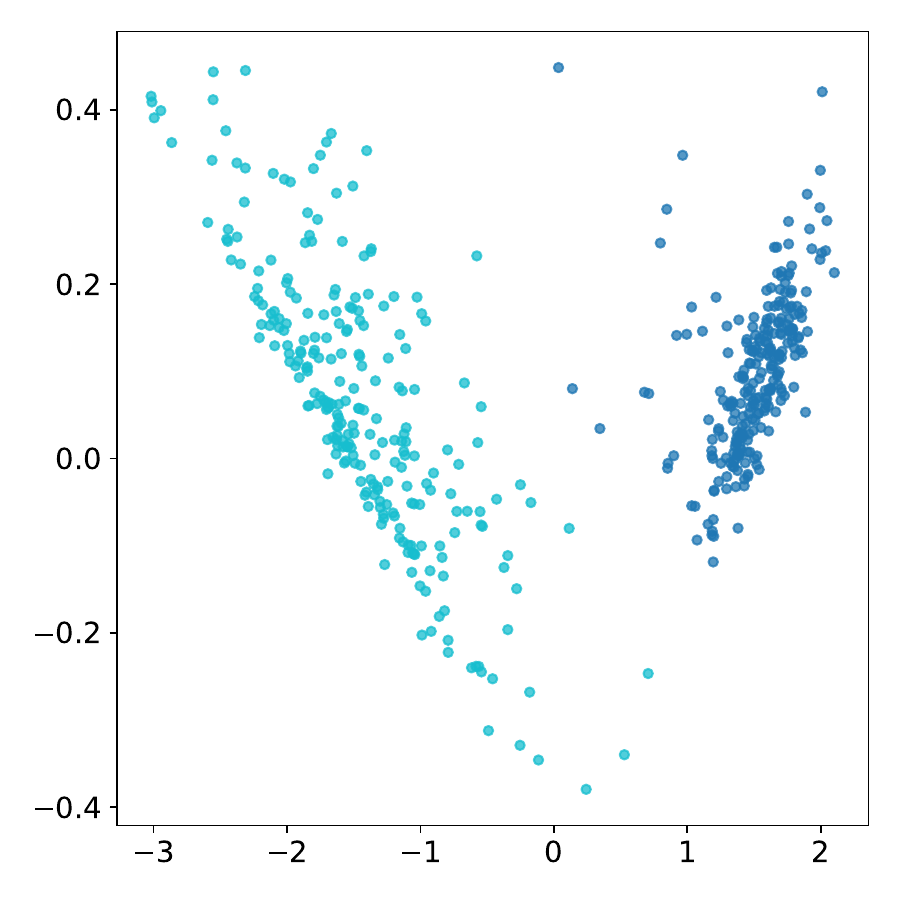}
    \caption{}
    \label{fig:before_removal}
  \end{subfigure}
  \begin{subfigure}{0.31\linewidth}
    \centering
    \includegraphics[width=1.0\linewidth]{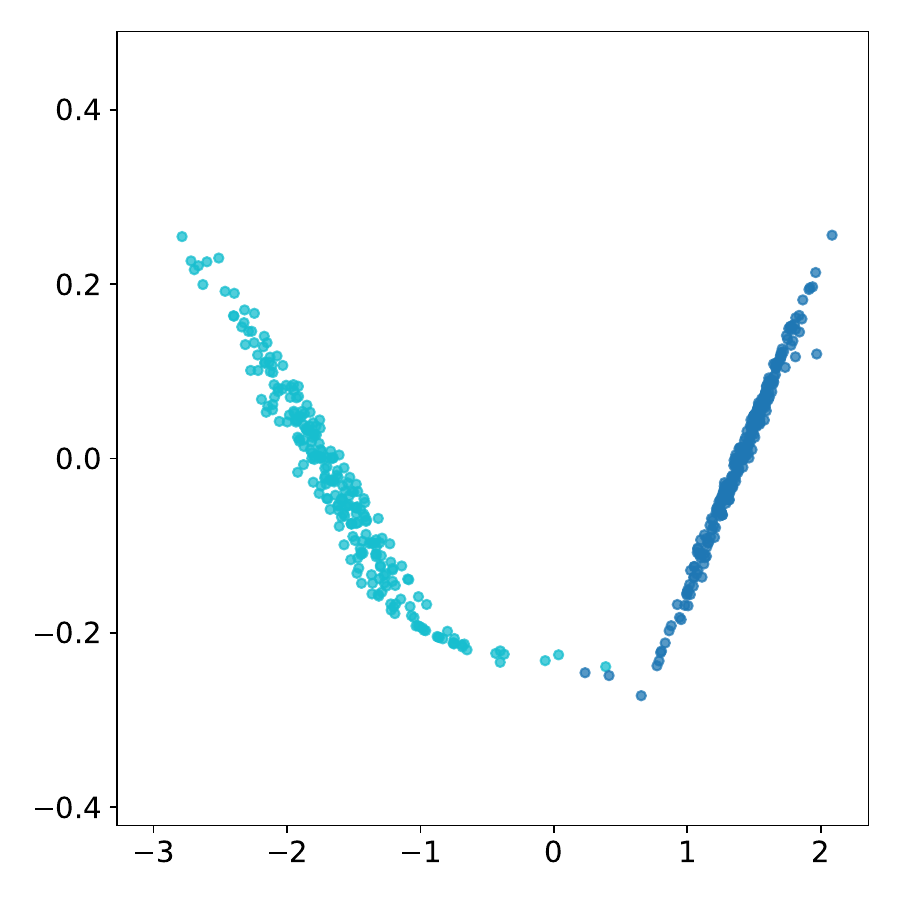}
    \caption{}
    \label{fig:after_removal}
  \end{subfigure}
  \caption{Effect of projection-removal in the two-dimensional feature space. (a) Input space depicting the decision boundaries. The solid line is the baseline classifier, and the dashed line is after projection removal training. Our method provides a wider, smoother margin. (b) Two‐dimensional PCA projection of the penultimate‐layer activations for the baseline model. (c) PCA projection of the same activations with projection removal training, exhibiting markedly tighter and more distinct class clusters. 
    % Illustration on a toy 2D Gaussian dataset showing the effect of projection removal in the feature space.  
    }
  \label{fig:short}
\end{figure*}

Since we enforce the correction jointly on clean and adversarial pairs during training, Lemma \ref{lem:complexity} predicts both improved clean generalisation and a tighter robust risk bound. The outcome is verified empirically in Section \ref{sec:experimnts}.

\subsection{Visual Illustration}  
To provide a clear demonstration of the effectiveness of our method, we employ a two-dimensional binary classification task based on a conditional Gaussian distribution. Each class is sampled from an isotropic Gaussian distribution with distinct means, creating a visually interpretable decision boundary. Here, we only consider the clean samples. The model is adversarially trained using PGD-10 attack.

Figure \ref{fig:projection_removal} overlays the learned boundaries. The solid boundary, obtained without projection removal, bends sharply and hugs the data. The dashed line, obtained with projection removal maintains a larger, more uniform margin. Projection removal during training noticeably changes the feature space. \ref{fig:before_removal} and \ref{fig:after_removal} show the plots of first two principal components of features from penultimate layer with and without projection removal training. Projection removal widens the gaps between classes in feature space. After using projection removal the leading components align with class-specific directions. Each class now occupies a subspace making their centroids farther apart and decision margins wider. Projection removal reallocates variance from tangled, inter-class axes to clean, intra-class axes, producing clear class separation in the penultimate layer. This reflects the theoretical reduction in Rademacher complexity as discussed in Lemma~\ref{lem:complexity}, and aligns with prior work that links flatter decision boundaries to better generalization and robustness~\cite{bartlett2002rademacher, neyshabur2017exploring}.

\section{Experiments}
\label{sec:experimnts}
This section presents a comprehensive evaluation of our proposed approach, \model. We begin by describing the experimental setup, including datasets, threat models, and implementation details. Next, we outline the baseline methods used for comparison. Finally, we present and analyze the results demonstrating the effectiveness of \model\ relative to state-of-the-art adversarial defenses.

\subsection{Experimental Setup}
\paragraph{Datasets} Our experiments focus on three commonly used benchmarks: CIFAR-10, CIFAR-100 \cite{krizhevsky2009learning}, SVHN \cite{netzer2011reading}, and \textcolor{black}{TinyImagenet} \cite{imagenet_cvpr09}.
% CIFAR-10 and CIFAR-100 use $32 \times 32$ pixel natural images to cover 10 and 100 unique classes, respectively. SVHN is a real-world digit identification dataset with images of house numbers (0-9) obtained from Google Street View. It also uses $32 \times 32$ resolution. It presents additional challenges due to background clutter, lighting variations, and other noise. These datasets are used as standard benchmarks for evaluating robust classification under adversarial settings.
\paragraph{Threat Model and Evaluation} Our evaluation uses the $\ell_{\infty}$ threat model. We set $\varepsilon = \frac{8}{255}$ for CIFAR-10, CIFAR-100 and SVHN, following standard parameters used in \cite{li2023squeeze}. To generate adversarial examples, we use PGD with 20 steps. We set step size $\alpha = \frac{2}{255}$ for all iterative attacks. In addition to PGD-based evaluations, we test robustness via the AutoAttack(AA) framework \cite{croce2020reliable}, which is widely recognized as a reliable robustness benchmark. We report the results for the checkpoint with best PGD-20 robust accuracy following \cite{zhang2020attacks, gowal2021improving, li2023squeeze}.

\paragraph{Implementation Details} To provide fair comparison, all methods are implemented using a consistent training procedure. Unless specified, models employ the ResNet-18 architecture as their backbone feature extractor, which was selected for its wide adoption and balanced complexity. To assess the scalability of our approach, we also conduct experiments with a larger-capacity WideResNet-34-10 architecture. Training is conducted for 120 epochs with stochastic gradient descent (SGD) optimizer, momentum of 0.9, weight decay fixed at $5 \times 10^{-4}$, and batch size set to 128. For \model\ specifically, the nearest-neighbor search is
performed within each batch. The projection removal coefficient $\lambda$ is fixed at 0.001 based on preliminary tuning experiments. We take $\beta$ as 6 for CIFAR-10 and SVHN and 4 for CIFAR-100. Notably, all hyperparameters, including attack configurations during training and evaluation, remains same as \cite{li2023squeeze}, across compared methods.

\begin{table*}[ht]
  \centering
    \caption{Clean and robust accuracies of adversarial training methods evaluated under the $\ell_\infty$ threat model with $\varepsilon=\tfrac{8}{255}$.  All models share the same ResNet-18 backbone and data pipeline. $^*$The authors have reported results for checkpoint that gives best sum of clean and AA accuracy.}

  \small
  \resizebox{0.7\textwidth}{!}{%
  \begin{tabular}{llcccccc}
    \toprule
    \multirow{2}{*}{Dataset} & \multirow{2}{*}{Method}
        & \multirow{2}{*}{Clean (\%)} & \multicolumn{5}{c}{Robust Accuracy (\%)} \\
        \cmidrule(lr){4-8}
         & & & FGSM & PGD-20 & PGD-100 & C\&W$_\infty$ & AA \\
     \midrule
    \multirow{9}{*}{CIFAR-10}
        & Vanilla AT            & 82.78 & 56.94 & 51.30 & 50.88 & 49.72 & 47.63 \\
        & TRADES                & 82.41 & 58.47 & 52.76 & 52.47 & 50.43 & 49.37 \\
        & MART                  & 80.70 & 58.91 & 54.02 & 53.58 & 49.35 & 47.49 \\
        & ST                    & 83.10 & \textbf{59.51} & 54.62 & 54.39 & 51.43 & \textbf{50.50} \\
        & SCARL                 &80.67 & 58.32 & 54.24 & 54.10 & \textbf{51.93} & 50.45 \\               
        & ARREST$^*$ & 86.63 & 57.70 & 49.40 & - & - & 46.14 \\ 
        & AR-AT$^*$              & \textbf{87.82} & - & 52.13 & - & - & 49.02 \\
        & DWL-SAT                    & 80.60 & - & 52.10 & - & 49.70 & 47.90 \\
        & \textbf{\model\ (ours)} & 81.26     & 59.37     & \textbf{54.82}     & \textbf{54.54}     & 50.07     & 49.14
        \\ \midrule
    \multirow{8}{*}{CIFAR-100}
        & Vanilla AT            & 57.27 & 31.81 & 28.66 & 28.49 & 26.89 & 24.60 \\
        & TRADES                & 57.94 & 32.37 & 29.25 & 29.10 & 25.88 & 24.71 \\
        & MART                  & 55.03 & 33.12 & 30.32 & 30.20 & 26.60 & 25.13 \\
        & ST                    & 58.44 & 33.35 & 30.53 & 30.39 & 26.70 & 25.61 \\
        & SCARL                 & 57.63 & 33.14 & 30.83 & 30.77 & 26.86 & 25.82 \\
        &AR-AT$^*$ & \textbf{67.51} & - & 26.79 & - & - & 23.38 \\
        & DWL-SAT                   & 56.70 & - & 29.00 & - & 26.90 & 23.90 \\
        & \textbf{\model\ (ours)} & 55.43     & \textbf{34.46}     & \textbf{31.55}     & \textbf{32.34}     & \textbf{28.19}     & \textbf{26.31}     \\ \midrule
    \multirow{6}{*}{SVHN}
        & Vanilla AT            & 89.21 & 59.81 & 51.18 & 50.35 & 48.39 & 45.96 \\
        & TRADES                & 90.20 & 66.40 & 54.49 & 54.18 & 52.09 & 49.51 \\
        & MART                  & 88.70 & 64.16 & 54.70 & 54.13 & 46.95 & 44.98 \\
        & ST                    & \textbf{90.68} & 66.68 & 56.35 & \textbf{56.00} & \textbf{52.57} & \textbf{50.54} \\
        & DWL-SAT                    & 89.80 & - & \textbf{57.30} & - & 51.70 & 46.10 \\
        & \textbf{\model\ (ours)} & 90.18     & \textbf{67.71}     & 56.61     & 55.64     & 50.20     & 48.35     \\
    \bottomrule
  \end{tabular}}
  % \caption{Clean and robust accuracies of adversarial training methods evaluated under the $\ell_\infty$ threat model with $\varepsilon=\tfrac{8}{255}$.  All models share the same ResNet-18 backbone and data pipeline. $^*$The authors have reported results for checkpoint that gives best sum of clean and AA accuracy.}
  \label{tab:main_results}
\end{table*}
 
\paragraph{Baselines} We benchmark \model\ against several state-of-the-art adversarial training methods. These baselines include: Vanilla Adversarial Training (Vanilla AT) \cite{madry2018towards}, uses PGD-based adversarial examples for robust model training. TRADES \cite{Zhang2019theoretically}, which explicitly trades off between robustness and accuracy via a tailored regularization term. MART \cite{Wang2020Improving}, which improves robustness by focusing on misclassified examples and integrating margin-based penalties. ST \cite{li2023squeeze} aims to tighten decision boundaries for better robustness. SCARL \cite{kuang2023scarl} introduces semantic information in model training by maximizing mutual information using text embeddings to improve adversarial robustness. ARREST \cite{satoshi2023adversarial} mitigates the accuracy–robustness trade-off by coupling adversarial finetuning with representation-guided knowledge distillation and noisy replay. AR-AT \cite{waseda2025rethinking}, introduces a one-sided invariance penalty that is applied exclusively to adversarial feature to improve clean accuracy. DWL-SAT \cite{xu2025dynamic} quantifies model robustness via robust distances and uses these distances to prioritize adversarial learning.

\subsection{Results}
\label{subsec:main_results}
% ================================================================
Table \ref{tab:main_results} reports the performance of all methods under identical training and attack settings. Across all three benchmarks, integrating \model\ into the MART backbone yields consistent improvements, and its advantages remain visible even when contrasted with the recent approaches. All results are reported under an $\ell_{\infty}$ threat model with $\varepsilon=8/255$. Baseline results are reported as in their original publications \cite{li2023squeeze, waseda2025rethinking, xu2025dynamic, satoshi2023adversarial}.

\paragraph{Evaluation on CIFAR-10}
\model\ improves robustness against FGSM attack to \textbf{59.37\,\%} and shows the highest robustness against PGD-20 and PGD-100 among all methods, recording \textbf{54.82\,\%} and \textbf{54.54\,\%} respectively. These scores improve on MART by $+0.46\,\%$, $+0.80\,\%$, and $+0.96\,\%$, respectively, while still exceeding ST by $+0.20\,\%$ (PGD-20) and $+0.15\,\%$ (PGD-100). Against the optimization based C\&W$_\infty$ attack, \model\ achieves \textbf{50.07\%}, surpassing both MART ($+0.72\%$) and DWL-SAT ($+0.37\%$). Robustness against AA increases to \textbf{49.14\,\%}, a $+1.65\,\%$ margin over MART, $+0.1.24\%$ over DWL-SAT, and $0.12,\%$ over the specialised AR-AT (49.02,\%). Projection removal filters gradient components that merely oscillate within the threat ball, allowing \model\ to focus on directions that truly threaten class boundaries. This selective suppression improves the worst case margins without perturbing the benign manifold.

\paragraph{Evaluation on CIFAR-100}
On the more granular 100 class task, \model\ raises PGD-20 robustness to \textbf{31.55\,\%}, improving on MART by $+1.23\,\%$, on ST by $+1.02\,\%$ and DWL-SAT by $+2.55,\%$. AA accuracy also increases to \textbf{26.31\,\%}, giving $+1.18\,\%$ over MART and $+0.70\,\%$ over ST, $+2.41,\%$ over DWL-SAT, and $+2.93,\%$ over AR-AT$)$. Clean performance remains competitive at \textbf{55.43\,\%} ($+0.40\,\%$ relative to MART).
% Here, projection removal aligns adversarial gradients with high–variance, class‐specific features, a property that grows increasingly valuable as the label space expands and inter class overlap deepens.

\paragraph{Evaluation on SVHN}
On the digit dataset \model\ delivers its significant relative benefits with clean accuracy increasing to \textbf{90.18\,\%} ($+1.48\,\%$ over MART and $+0.38\%$ over DWL-SAT), and PGD-20 robustness reaches \textbf{56.61\,\%}, surpassing MART by $+1.91\,\%$ and slightly improving over ST by $+0.26\,\%$. SVHN images have relatively simple backgrounds and well separated digit classes, which leads to lower inter-class ambiguity in the feature space. Consequently, the scope for improvement from nearest-neighbor projection removal is more limited than on more complex datasets.
% Projection removal supresses the high–frequency artefacts that dominate adversarial perturbations on low–resolution imagery.

\subsection{Scalability to Larger Architecture and Datasets} 
To further verify that \model\ generalises beyond small backbones, we repeat the evaluation on WideResNet-34-10 (WRN-34-10). Table~\ref{tab:wideresnetcifar10trades} reports clean and robust accuracies on CIFAR-10. On WRN-34-10, \model\ attains the highest robust accuracy of 58.40\% against PGD$_{TRADES}$ \cite{Zhang2019theoretically} improving on ST by +0.67\% and on TRADES by +1.75\%.  The AA performance (51.33\%) also stays competitive, exceeding MART. These results indicate that projection removal continues to tighten decision boundaries even as model capacity grows, yielding a net gain against strong attacks without compromising benign accuracy. Similar to the ResNet-18 case, the advantage of \model\ is most visible under iterative attacks. While ST excels on AA, \model\ provides the best defence against 20-step PGD. The geometric regularisation imposed by projection removal helps WRN-34-10 avoid the over-fitting to specific attack patterns that has been reported for wider networks \cite{rice2020overfitting}.

\textcolor{black}{We also evaluate our approach on TinyImageNet. The proposed method strengthens robustness across both backbones while keeping benign accuracy within a comparable operating range to established defences. On WRN-34-10, it attains 26.53\% under PGD-20, improving over ST by +1.29 percentage points and over TRADES by +3.20 points, and closely tracking the strongest reported baseline. On ResNet-18, it delivers the top PGD-20 score at 13.04\%, exceeding MART by +0.46 points and ST by +1.37 points.}

\textcolor{black}{Overall, the experiment confirms that \model\ scales gracefully, maintaining or improving robustness compared with state-of-the-art training objectives even on large-capacity architectures and datasets.}

\begin{table}[t]
  \centering
    \caption{WRN-34-10 on CIFAR-10 (\(\ell_\infty,\varepsilon=\tfrac{8}{255}\)). Robust accuracy is measured against PGD$_{TRADES}$ \cite{Zhang2019theoretically} and AA. }

  \small
  \resizebox{0.8\columnwidth}{!}{%
  \begin{tabular}{lccc}
    \toprule
    Method & Clean (\%) & PGD-20 (\%) & AA (\%)\\
    \midrule
    TRADES            & 84.80 & 56.65 & 52.94\\
    MART              & 84.17 & —     & 51.10\\
    ST                & \textbf{84.92} & 57.73 & \textbf{53.54}\\
    % ARREST            & 90.24 & 52.40 & 50.20 \\
    % AR-AT            & 91.2 & 52.24 & 50.77 \\
    \model\      & 83.53 & \textbf{58.40} & 51.33\\
    \bottomrule
  \end{tabular}
  }
  % \caption{WRN-34-10 on CIFAR-10 (\(\ell_\infty,\varepsilon=\tfrac{8}{255}\)). Robust accuracy is measured against PGD$_{TRADES}$ \cite{Zhang2019theoretically} and AA. }
  \label{tab:wideresnetcifar10trades}
\end{table}

\begin{table}[t]
  \centering
  \small
    \caption{Comparison on TinyImagenet (\(\ell_\infty,\varepsilon=\tfrac{8}{255}\)). Robust accuracy is measured against PGD-20.}

  \resizebox{\columnwidth}{!}{%
  \begin{tabular}{lcccc}
    \toprule
    \multirow{2}{*}{Method} 
    & \multicolumn{2}{c}{\textcolor{black}{WRN-34-10}} 
    & \multicolumn{2}{c}{\textcolor{black}{ResNet-18}} \\
    \cmidrule(lr){2-3} \cmidrule(lr){4-5}
    & Clean (\%) & PGD-20 (\%) & Clean (\%) & PGD-20 (\%) \\
    \midrule
    TRADES         & \textbf{49.22} & 23.33 & -     & -     \\
    MART           &   46.94   &   \textbf{26.82}   & 27.56 & 12.58 \\
    ST             &   47.97   &   25.24   & \textbf{29.35} & 11.67 \\
    \model         & 42.71 & 26.53 & 27.43 & \textbf{13.04} \\
    \bottomrule
  \end{tabular}%
  }
  % \caption{Comparison on TinyImagenet (\(\ell_\infty,\varepsilon=\tfrac{8}{255}\)). Robust accuracy is measured against PGD-20.}
  \label{tab:tinyimagenet}
\end{table}

\subsection{Ablation Study}
\begin{figure}[htbp]
    \centering
    \begin{subfigure}[b]{0.49\columnwidth}
        \includegraphics[width=\linewidth]{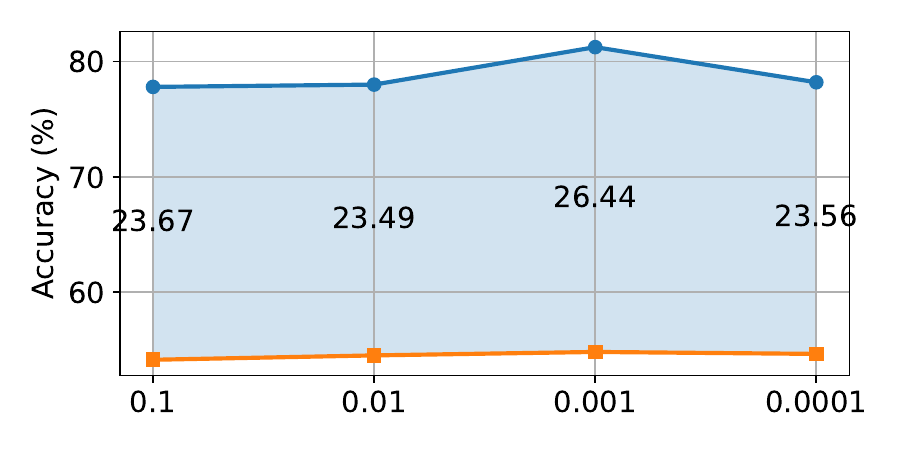}
        \caption{$\lambda$}
        \label{fig:ablation_lambda}
    \end{subfigure}
    \hfill
    \begin{subfigure}[b]{0.49\columnwidth}
        \includegraphics[width=\linewidth]{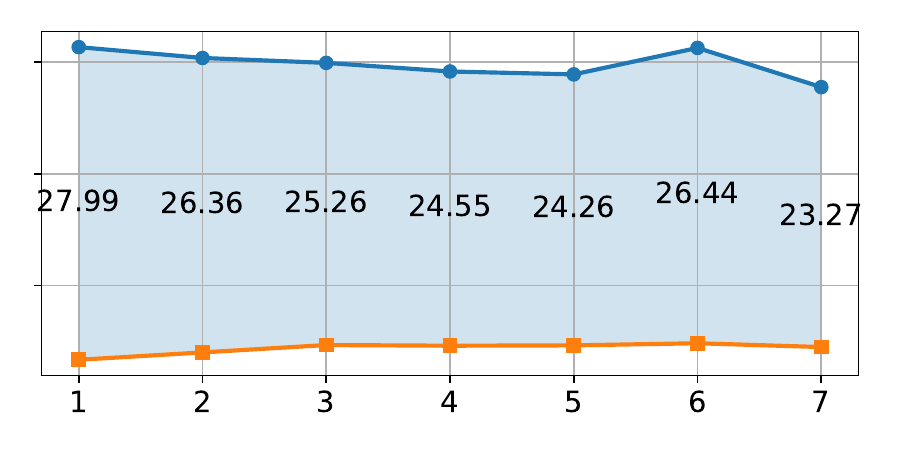}
        \caption{$\beta$}
        \label{fig:ablation_beta}
    \end{subfigure}
    \caption{Clean (circle) and robust (square) accuracy under different (a)$\lambda$ and (b)$\beta$ values.  Shaded areas show the clean–robust gap.}
    \label{fig:lambda_beta}
\end{figure}

We evaluate two hyperparameters for ResNet-18 on CIFAR-10, projection removal strength $\lambda$ and regularization weight $\beta$, which scales the regularizer. Figures \ref{fig:ablation_lambda} and \ref{fig:ablation_beta} plot clean and robust accuracy under different settings.

\paragraph{Projection Removal Strength ($\lambda$)}
We vary $\lambda\in{0.1, 0.01, 0.001, 0.0001}$ keeping $\beta=6$.  At $\lambda=0.001$, clean accuracy peaks at 81.26\% while robust accuracy reaches 54.82\%. Both metrics drop by roughly 2\% when $\lambda$ is an order of magnitude higher or lower. Projection removal raises robust accuracy, yet different values of $\lambda$ change it only slightly (54.14–54.82 \%). Clean accuracy, however, varies much more.

\paragraph{Regularization Weight ($\beta$)}
We vary $\beta\in{1,2,3,4,5,6,7}$ with $\lambda=0.001$. As shown in Figure \ref{fig:ablation_beta}, clean and robust accuracy both vary by only a small margin across this range.  The stability of both metrics indicates that scaling the regularizer alone has minimal impact on the model accuracy.

\begin{figure}
\centering
\includegraphics[width=0.98\linewidth]{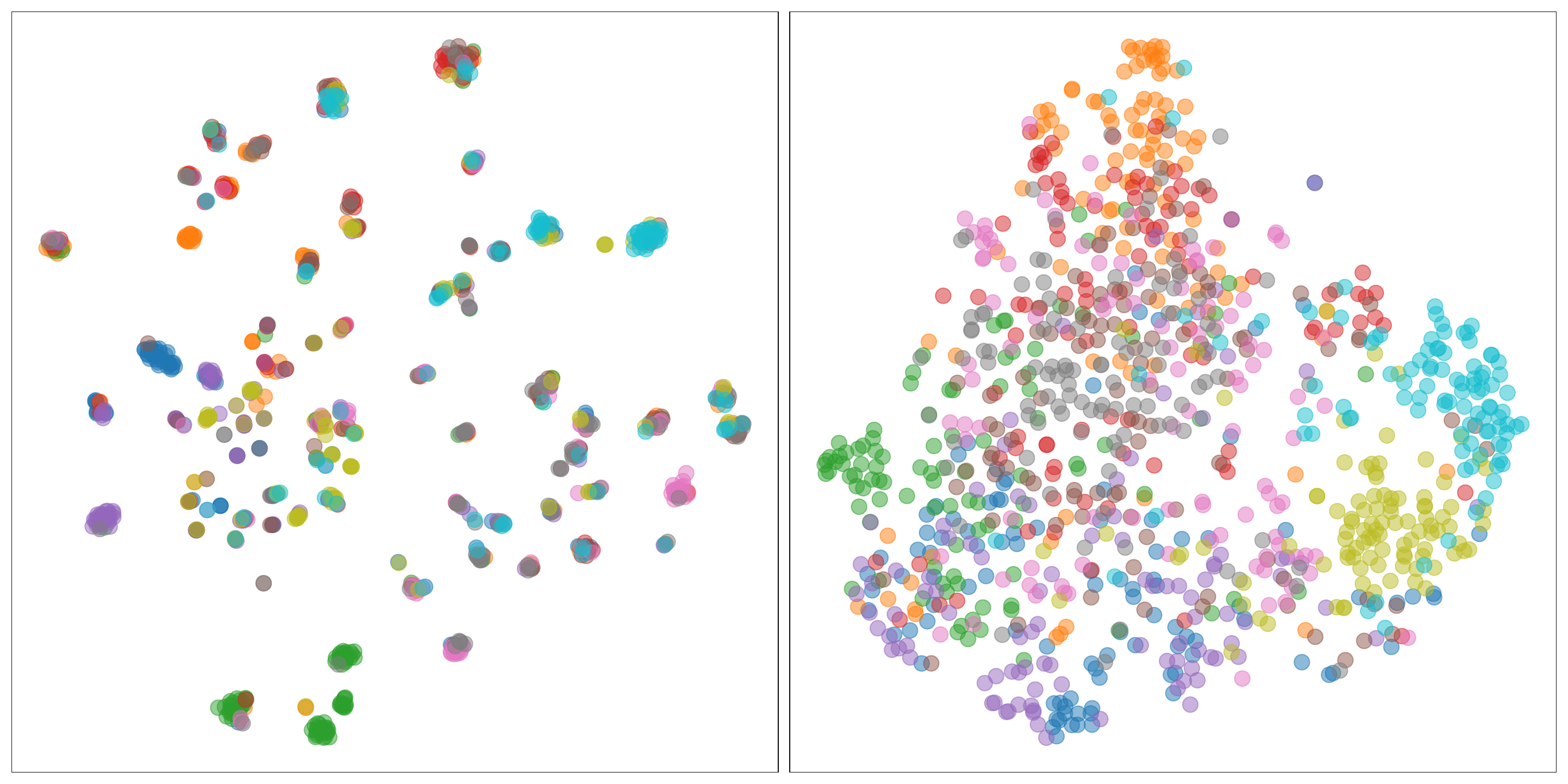}%
\caption{t-SNE visualization of CIFAR-100 on ResNet18 without projection removal training (left) and with projection removal training (right).}
\label{fig:feature_visualisation}
\end{figure}

\textcolor{black}{\paragraph{Feature Space}
We further visualize adversarial features with t-SNE \cite{vandermaaten08a}. We extract features of 10 random classes of PGD-attacked CIFAR-100 samples from Resnet-18 model with and without projection removal training and embed them with t-SNE. As illustrated in figure \ref{fig:feature_visualisation}, \model\ yields few, large, contiguous class clusters, while without projection removal training the classes spread across multiple interleaved clusters. The clearer, less fragmented clusters of \model\ indicate stronger neighborhood preservation and reduced manifold shattering under attack, indicating its  robustness. Following \cite{pmlr-v157-kallidromitis21a, Nie2008TraceRC} we also report Fisher \cite{https://doi.org/10.1111/j.1469-1809.1936.tb02137.x} and silhouette scores \cite{ROUSSEEUW198753}. Fisher score compares between-class spread to within-class scatter, larger values indicate better class separability. Similarly, silhouette score contrasts average distance of a point to its own class with that to the nearest other class. A negative silhouette score means that, on average, a point is closer to another class cluster than to its own, it is likely misassigned or classes are overlapping. The Resnet-18 trained with projection removal attains a higher silhouette score (0.009 vs -0.008) and a higher Fisher ratio  (0.46 vs 0.13), confirming stronger class separation under attack and supporting the robustness of \model.}

% \begin{figure}
% \centering
% \includegraphics[width=\linewidth]{wacv2026/figures/beta.pdf}
% \caption{Clean (circle) and robust (square) accuracy under different $\beta$  values. Shaded areas show the clean–robust gap.}
% \label{fig:ablation_beta}
% \end{figure}

% \begin{figure}[t]
% \centering
% \begin{subfigure}[t]{0.48\linewidth}
%     \centering
%     \includegraphics[width=\linewidth]{wacv2026/figures/lambda.pdf}
%     \caption{Accuracy under different $\lambda$ values.}
%     \label{fig:ablation_lambda}
% \end{subfigure}%
% \hfill
% \begin{subfigure}[t]{0.48\linewidth}
%     \centering
%     \includegraphics[width=\linewidth]{wacv2026/figures/beta.pdf}
%     \caption{Accuracy under different $\beta$ values.}
%     \label{fig:ablation_beta}
% \end{subfigure}
% \caption{Clean (circle) and robust (square) accuracy under varying hyperparameters. Shaded areas indicate the clean–robust gap.}
% \label{fig:ablation}
% \end{figure}

\section{Conclusion}
\label{sec:conclusion}
% Projection removal widens the decision boundary only where it overlaps with the nearest inter-class features. It reduces the intra-class variance. This adjustment yields consistent gains against strong white-box attacks while preserving benign accuracy. The gains are even larger on CIFAR-100, which has a wider label space; here, \model\ achieves the highest accuracy across all attacks. These improvements arise despite using identical optimizer schedules and attack hyper-parameters. We also theoretically show that our method reduces the model complexity which helps in generalization.

Our projection removal method widens the decision boundary only along locally vulnerable directions where a sample aligns with its nearest inter-class features. Unlike prior feature-space regularization methods that impose global geometric constraints or modify the inner maximization, \model\ applies a sample-conditioned correction by subtracting the feature component aligned with the nearest impostor direction. This targeted operation reduces inter-class entanglement while preserving intra-class structure, leading to consistent gains against strong white-box attacks without sacrificing benign accuracy. The improvements are most pronounced on CIFAR-100, where \model\ achieves the strongest robustness across evaluated attacks. These gains are obtained with identical optimizer schedules and attack hyper-parameters, and are supported by our theoretical analysis showing reduced model complexity and improved generalization.

\section{Acknowledgement}
This work was supported in part by the iHUB-ANUBHUTI-IIITD Foundation, established under the NM-ICPS scheme of the Department of Science and Technology, Government of India, and in part by the Anusandhan National Research Foundation (ANRF), Department of Science and Technology, Government of India (Project No. CRG/2022/004069).

\bibliographystyle{IEEEtran}
\bibliography{main}

\end{document}